\documentclass{article} 
\usepackage{iclr2024_conference,times}

\usepackage{amsmath,amsfonts,bm}

\def\eqref#1{equation~\ref{#1}}

\def\1{\bm{1}}

\DeclareMathAlphabet{\mathsfit}{\encodingdefault}{\sfdefault}{m}{sl}
\SetMathAlphabet{\mathsfit}{bold}{\encodingdefault}{\sfdefault}{bx}{n}

\usepackage{hyperref}
\usepackage{url}

\usepackage{algorithm}
\usepackage{algpseudocode}
\usepackage{amsfonts}       
\usepackage{amssymb,amsmath,amsthm}
\usepackage{array}
\usepackage{bm}
\usepackage{booktabs}       
\usepackage{cleveref}
\usepackage{dcolumn}
\usepackage{enumitem}
\usepackage[T1]{fontenc}    
\usepackage{graphicx}
\usepackage[utf8]{inputenc} 
\usepackage{lineno}
\usepackage{nicefrac}       
\usepackage{mathtools}
\usepackage{microtype}      
\usepackage{placeins}
\usepackage{setspace}
\usepackage{subcaption}
\usepackage{tabularx}
\usepackage{tcolorbox}
\usepackage{url}            
\usepackage{wasysym}
\usepackage{wrapfig}
\usepackage{xcolor}         
\usepackage{xspace}
\newcommand{\openabm}{OpenABM-Covid19}
\newcommand{\covidn}{COVID19}

\newcommand{\covidscore}{\textsc{covidscore}}
\newcommand{\Reals}{\mathbb{R}}
\newcommand{\norm}[1]{\,\left\Vert {#1} \right\Vert\,}

\newtheorem{lemm}{Lemma}
\newtheorem{theorem}{Theorem}

\definecolor{blue}{RGB}{66,133,244}

\newcommand{\veryshortarrow}[1][3pt]{\mathrel{%
   \hbox{\rule[\dimexpr\fontdimen22\textfont2-.2pt\relax]{#1}{.4pt}}%
   \mkern-4mu\hbox{\usefont{U}{lasy}{m}{n}\symbol{41}}}}

\title{DNA: Differentially private Neural \\ Augmentation for contact tracing}

\author{
Rob Romijnders\textsuperscript{\rm 1}, 
Christos Louizos\textsuperscript{\rm 2}, 
Yuki M. Asano\textsuperscript{\rm 1}, 
Max Welling\textsuperscript{\rm 1}  \\
\textsuperscript{\rm 1}University of Amsterdam \ \textsuperscript{\rm 2}Qualcomm AI research}

\iclrfinalcopy 
\begin{document}

\maketitle

\begin{abstract}
The \covidn\ pandemic had enormous economic and societal consequences. 
Contact tracing is an effective way to reduce infection rates by detecting potential virus carriers early.
However, this was not generally adopted in the recent pandemic, and privacy concerns are cited as the most important reason.
We substantially improve the privacy guarantees of the current state of the art in decentralized contact tracing. 
Whereas previous work was based on statistical inference only, we augment the inference with a learned neural network and ensure that this neural augmentation satisfies differential privacy. 
In a simulator for \covidn\, even at $\varepsilon=1$ per message, this can significantly improve the detection of potentially infected individuals and, as a result of targeted testing, reduce infection rates.
This work marks an important first step in integrating deep learning into contact tracing while maintaining essential privacy guarantees.
\end{abstract}

\section{Introduction} \label{sec:introduction}

The \covidn\ pandemic had enormous consequences~\citep{econ_impact_01,econ_impact_02,societal_impact_01,societal_impact_02}. 
Contact-tracing algorithms could make early predictions of virus carriers, signaling individuals to get tested and thereby reducing the spread of the virus~\citep{baker2021epidemic}.
However, population surveys show that privacy concerns are among the primary reasons for the low adoption of these algorithms~\citep{privacy_concern_01,privacy_concern_02,privacy_concern_03}.

Recent work introduces a differential privacy (DP) solution against a privacy attack on contact tracing algorithms~\citep{romijnders2024protectyourscore}. This method, however, uses only a statistical model with a corresponding inference method to make predictions of infectiousness~\citep{DBLP:conf/uai/Rosen-ZviJY05}. We propose augmenting the inference steps with a learned neural network. Such a `neural augmentation' can learn patterns in the data that are not captured by the statistical model. This builds on a line of work about augmenting statistical updates with learnable functions~\citep{neural_augment_01,neural_augment_02,gregor2010learning}. We ensure differential privacy for the neural augmentation and name the method Differentially private Neural Augmentation (DNA).

In the attack scenario, an adversary tries to infer the private state of a victim. The model predicts for every user on every day a risk score, named \covidscore, and it is this private score that the attacker tries to infer. We quote from previous work~\citep{romijnders2024protectyourscore}: 

\begin{quote}
`An adversary wants to determine the \covidscore\ of a victim. The adversary installs the app and only makes contact with the victim. The next day, the adversary observes a change in their  \covidscore. This change is due to the victim, and the adversary reconstructs the \covidscore\ of the victim.'
\end{quote}

Our goal is to achieve better predictions of the \covidscore\ for an algorithm that is DP against this attack. Injecting noise is the default method for guaranteeing DP, but the noise makes the predictions worse. In order to motivate neural augmentation, we identify a hierarchy of methods to achieve DP. Whereas previous work analyzed contact tracing in terms of individual messages per user (level 1) and multiple messages per day (level 2), our analysis considers the set of all messages in a window of multiple days (level 3). 
The analysis reveals that the current algorithm has a bounded sensitivity, which motivates us to propose a neural augmentation module with a similar bound on the sensitivity.

\newpage

In total, we make the following contributions: 

\begin{itemize}
\item We show how statistical inference can be augmented with a neural network such that the combined prediction satisfies differential privacy. We are the first to bridge the promising field of neural augmentation with differential privacy.
\item We identify a novel hierarchy of privacy in contact tracing and provide a theoretical proof of differential privacy at the most general level of the hierarchy. 
\item Both methods are tested on a widely used simulator. Even in the challenging situation of noisy tests, or agents not following the protocol, our method significantly reduces the number of simultaneously infected individuals, which is a key marker for pandemic mitigation.
\end{itemize}

The code for running the experiments can be found at \href{https://github.com/RobRomijnders/DNA}{github.com/RobRomijnders/DNA}.

\section{Related work}\label{sec:related_work}

As this work brings together multiple fields, we overview related literature in four areas: differential privacy, neural augmentation, Lipschitz-constrained neural networks, and the application area of statistical contact tracing.

We use differential privacy (DP) as the main method to quantify the privacy of a statistical algorithm. A good overview of DP is the book by~\citet{dwork_dp}. Differential privacy has been used in fields such as mean estimation~\citep{dp_mean_estimation}, deep learning~\citep{abadi_dpsgd}, and statistical query answering~\citep{goryczka2015comprehensive}. Moreover, DP is named as a promising area for federated learning research~\citep{kairouz2021advances}. Another reason for using DP is the use of building blocks like the post-processing property~\citep{dwork_dp} and advanced composition theorems~\citep{advanced_composition_dp,abadi_dpsgd,ponomareva2023dp}. 

Neural augmentation is a method for augmenting statistical/physical algorithms with learnable neural networks. This has been studied in application areas such as sparse coding~\citep{gregor2010learning}, MRI reconstruction~\citep{neural_augment_02}, and error correction codes~\citep{neural_augment_01}. We are the first to combine neural augmentation and differential privacy. 

For the DP guarantee, we will make use of neural networks with constrained Lipschitz constant for the input data (defined in Section~\ref{sec:dp_neural_augment}). The concept of Lipschitz neural networks has been studied as a method in adversarial robustness~\citep{spectral_norm_adv_robustness}, and to optimize a constrained family of models in Generative Adversarial Networks~\citep{goodfellow2014generative,miyato2018spectral}. Previous works use Lipschitz-constrained models for differentially private learning~\citep{chaudhuri2011differentially,lipschitz_way,minami2016differential} and data analysis~\citep{jha2013testing}. In our case, we need a Lipschitz constraint with respect to the input data, and we follow previous work~\citep{jha2013testing} to apply this to neural augmentation.

Contact tracing has been shown to be effective in mitigating a pandemic outbreak~\citep{jenniskens2021effectiveness}. Previous work shows that models based on statistical algorithms can outperform traditional methods~\citep{baker2021epidemic,crisp}. However, population studies make clear that privacy concerns are a major reason for not using contact tracing algorithms~\citep{privacy_concern_01,privacy_concern_02,privacy_concern_03}. A recent work introduced a DP alternative for statistical contact tracing~\citep{romijnders2024protectyourscore}. That work, however, a) considers only a privacy composition per day which is only level 2 of our hierarchy and b) uses only classical statistical updates. In this work, we propose a DP method in a higher level of privacy hierarchy and show that, combined with neural augmentation, this can substantially decrease the peak impact of the infection.

\section{Method}\label{sec:method}

We provide a background on the statistical model and prove differential privacy for each level of the hierarchy. Appendix~\ref{app:notation} provides an overview of the notation used in this paper.

\subsection{Statistical model}

We will formulate a common statistical model for contact tracing. A Markov chain of random variables $z_{u,t}$ models the disease progression for a user $u$ at timestep $t$. Each such variable takes on one of four disease states, ${S,E,I,R}$, for the Susceptible, Exposed, Infected, and Recovered state~\citep{kermack1927contribution,anderson1992infectious}. The dynamics of this Markov Chain are described in Equation~\ref{eqn:seir_dynamics_full}. The only non-scalar transition function is the transition between state S and state E:

\begin{align}\label{eqn:crisp_noisy_or}
   P(z_{u,t+1} = E|z_{u,t}=S, z_{N(u,t)}) = 1 - (1-p_0) (1-p_1)^{|\{z_c \ \in \ z_{N(u, t)}: \ z_{c}=I \}|}
\end{align}

Parameter $p_1$ indicates the probability of transmitting the virus upon contact, and its value is set to a value from previous literature~\citep{romijnders2023notimetowaste,crisp}. A user can have contact with multiple other users on a particular day. As such, $z_{N(u,t)}$ denotes the set of random variables of all contacts of user $u$ at time step $t$. 

The actual observation is a test for \covidn\, which can have a false positive or false negative outcome regarding the underlying state. The data set of observations is $D_\mathcal{O} = \{o_{u_i,t_i}  \}_{i=1}^O$, which are $O$ observations, each with an outcome $\{0,1\}$ for user $u_i$ at time step $t_i$. The observation model follows previous literature~\citep{crisp} and is stated in Equation~\ref{eqn:test_outcome}. We denote the False Positive rate (FPR) by $\beta$ and denote the False Negative rate (FNR) by $\alpha$. 

\subsection{Statistical messages}

The inference for the statistical SEIR model consists of sending decentralized messages between users. We follow the algorithm of~\citep{romijnders2024protectyourscore} and run the Factorised Neighbors (FN) algorithm~\citep{DBLP:conf/uai/Rosen-ZviJY05}. This algorithm has been shown to be effective for decentralized contact tracing and its updates are amenable to the privacy analysis that we make in this paper. 

We can write the FN inference algorithm as a function of the incoming messages.  Each message is a number in the range $[0,1)$ and denotes the \covidscore, $\phi_{c,t}$, of user $c$ at timestep $t$.  On each day $t$, user $u$ has $C_t$ contacts with a user that will be denoted by the relative index $1, 2, \cdots, C_t$. 

\subsection{Differential privacy} \label{sec:dp_hierarchy}

The conventional definition of differential privacy is used, $(\varepsilon,\delta)$-DP~\citep{dwork_dp}. For every $\varepsilon > 0$, $\delta \in [0, 1]$, a function $f(\cdot)$, for any outcome $\Phi$ in the range of $f(\cdot)$, and two adjacent data sets $D$, $D'$ that have at most one element different, the following constraint holds:

\begin{align}
    p(f(D) \in \Phi) \leq e^\varepsilon p(f(D') \in \Phi) + \delta \label{eqn:definition_dp}
\end{align}

We define two data sets, $D, D'$, as adjacent when the \covidscore\ of one of the contacts has a different value. A dataset $D$ is a collection of \covidscore\ that $C$ contacts send at particular days, $D = \{ (\mu_i, t_i) \}_{i=1}^C$. Each $t_i \in \{ 1, 2,  \cdots, T-1\}$ is a particular day when a contact occurs, where $T$ is the window length. The values for $t_i$ are public information from the point of view of the attack model, i.e. an attacker could know on what day a particular contact was established, but the \covidscore\ of that contact should remain private. Therefore, DP is defined between two adjacent datasets, where the value of one message by a contact, $\mu_i$, is replaced by another value $\mu'_i$. The largest change in the output of an algorithm then defines the sensitivity:

\begin{equation} \label{eqn:sensitivity_definition}
\Delta = \max_{\mu_1, \mu'_1 } \norm{f\big( \ \{(\mu_1, t_1)\} \cup D \ \big) - f\big( \ \{(\mu'_1, t_1)\} \cup D \ \big)} \qquad \forall \ D. 
\end{equation}

For a particular sensitivity, the (Analytical) Gaussian mechanism~\citep{dwork_dp,balle2018gauss} prescribes the standard deviation of additive Gaussian noise such that the output of the algorithm satisfies $(\varepsilon,\delta)$-DP.

\begin{wrapfigure}{r}{0.4\textwidth}
  \begin{center}
    \includegraphics[width=0.4\textwidth]{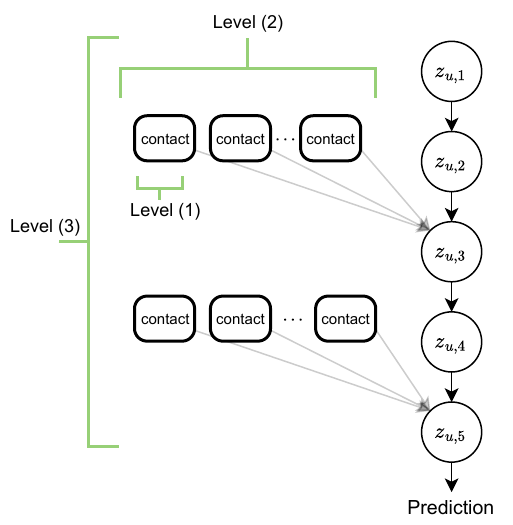}
  \end{center}
  \caption{Three levels of DP analysis.}
  \vspace{-5mm}
\end{wrapfigure}
\textbf{DP analysis per message (level 1): } If each individual message is noised, by the post-processing property of DP, the entire prediction function is DP. We add Gaussian noise according to the Gaussian mechanism to ascertain DP~\citep{dwork_dp,romijnders2024protectyourscore}.

\textbf{DP analysis of messages per day (level 2): } The statistical model deals with messages per day in a product, highlighted in Equation~\ref{eqn:crisp_dynamics_under_exp}. Previous work suggests adding log-normal noise to each message and provides the corresponding DP analysis~\citep{romijnders2024protectyourscore}.

\textbf{DP analysis of messages per multiple days (level 3): } A novel contribution of this work is that we generalize the hierarchy of privacy on one more level. We establish a bound on the sensitivity of the prediction function in the presence of multiple contacts on multiple days. Based on the sensitivity, we add Gaussian noise according to the Gaussian mechanism to ensure DP. The sensitivity of FN is formalized in the following theorem.

\begin{theorem} \label{thm:sensitivity_14day}
    Following notation in Section~\ref{sec:method} and assuming that each message $\mu_{i}$ is bounded in the interval $[0, \gamma_u]$, for any two adjacent datasets as defined in Equation~\ref{eqn:sensitivity_definition}, the sensitivity for the FN inference function, $F(\cdot)$, is defined by:
    \begin{equation}
        \Delta = \max_{\mu_1, \mu'_1  \in [0, \gamma_u]} \norm{F\big( \ \{(\mu_1, t_1)\} \cup D \ \big) - F\big( \ \{ (\mu'_1, t_1)\} \cup D \ \big)} \leq p_1 \gamma_u \qquad \forall \ D. 
    \end{equation}
\end{theorem}

\textit{Proof sketch: } Each contact has a probability $p_1$ of transmitting the virus during a contact, c.f. Equation~\ref{eqn:crisp_noisy_or}, and the value of the message is in the range $0$ to $\gamma_u$. For every sequence of health states $S \veryshortarrow E \veryshortarrow I \veryshortarrow R$, the probability can only increase $p_1$ for a message value up to $\gamma_u$, of which the output is a convex combination. The sensitivity of $p_1 \gamma_u$ follows. See Section~\ref{app:global_sens} for the full proof. \qed.

This method modifies the DPFN algorithm based on bounding the sensitivity. Therefore, the experimental results will refer to this method as DPFN-sensitivity, or DPFN-S for short.

\section{Neural Augmentation} \label{sec:dp_neural_augment}

We propose to improve the statistical predictions with neural augmentation. The current SEIR model has only four modifiable parameters and we hypothesize that a neural network can learn more complex patterns from data~\citep{neural_augment_01,neural_augment_02}. Theorem~\ref{thm:sensitivity_14day} shows that the FN function has a sensitivity of $p_1$, typically around 0.02~\citep{openabm}. However, such a sensitivity is only 2\% of the output range, which is $\phi_{u,t} \in [0,1]$. This shows that even a function with relatively low sensitivity can predict infectiousness. Therefore, we aim to learn a neural augmentation with similarly low sensitivity to augment the statistical predictions while maintaining privacy.

To obtain a bound for the sensitivity of a neural network, we use Lipschitz-constrained neural networks ~\citep{lipschitz_way}. A function $g:\Reals^m\rightarrow\Reals^n$ has Lipschitz constant $l$  if for every $x,y\in\Reals^m$ we have $\|g(x)-g(y)\|_2 \leq l\|x-y\|_2$. This constraint is achieved when the gradient norm is upper bounded, $\sup_x \norm{ \nabla_x g(x)} \leq l $~\citep{lipschitz_way}.

The Lipschitz constant can be decomposed for a neural network into the Lipschitz constants of its layers. For a function $g(x) = g_1 \circ g_2 \circ \cdots \circ g_H (x)$ based on $H$ composite functions, denote by $l_h$ the Lipschitz constant of layer $h$. Then the Lipschitz constant of the composite function, $g(\cdot)$, is: 

\begin{equation}\label{eqn:lipschitz_constant}
l = l_1 \times l_2 \times \cdots \times l_H. 
\end{equation}

For linear layers in the neural network, the Lipschitz constant equals the spectral norm, which will be restricted during training~\citep{miyato2018spectral,bartlett2017spectrally}. We also use an activation function with a bounded gradient, such as the Rectified Linear Unit~\citep{relu}.
    
The neural network takes, as features, the messages that are sent to the agent by its contacts and predicts the infectiousness for that particular agent on that day. As there can be a variable number of messages and the prediction is invariant under permutation, we use a  DeepSet model~\citep{deepset}. For each contact, we denote the stacked feature vector $x_i = [\mu_i,  t_i]^T$.  Each feature vector is mapped to a representation by a neural network $g^{(1)}(\cdot)$, and, after averaging, a second neural network makes the prediction, $g^{(2)}(\cdot)$. The total neural network model is as follows:

\begin{align}
    \phi &= G_\theta(\{( \mu_i, t_i )\}_{i=1}^{C_T}) = g^{(2)}_\theta( \quad \frac{1}{C} \sum_i g^{(1)}_\theta([\mu_i, t_i]^T) \quad ). \label{eqn:neural_network}
\end{align}

\begin{theorem} \label{thm:lip_mlp}
    The neural network in Equation~\ref{eqn:neural_network} has Lipschitz constant $\frac{1}{C}$ w.r.t. one vector $[\mu_i, t_i]^T$.
\end{theorem}
\textit{Proof: } The mean function multiplies each vector, $g^{(1)}_\theta([\mu_i, t_i]^T)$, from a single message $\mu_i$ with $\frac{1}{C}$. The Lipschitz constant of all linear layers and activation functions in $g^{(1)}_\theta$ and $g^{(2)}_\theta$ does not exceed 1. Therefore, by Equation~\ref{eqn:lipschitz_constant}, the product of Lipschitz constants does not exceed $\frac{1}{C}$ \qed

Algorithm~\ref{algo:mlp_lipschitz} denotes the full algorithm for use in contact tracing.  With the sensitivity of the statistical prediction and neural augmentation together, we add noise according to the Gaussian mechanism to ensure $(\varepsilon,\delta)$-DP. The crucial differences with the DPFN-sensitivity method are indicated in blue.

\begin{algorithm}
\setstretch{1.45}
\caption{DNA: differential private neural augmentation}
\begin{algorithmic}
\Require Dataset $D = \{( \mu_i, t_i )\}_{i=1}^{C_t}$, constants $p_1, \gamma_u \in (0,1)$, DP $(\varepsilon, \delta)$
\Ensure output $\phi$ is differentially private with budget $(\varepsilon, \delta)$
\State $\mu_i \gets \min(\mu_i, \gamma_u)$  \Comment{Clip message value}
\State $\xi \sim \mathcal{N}\big(0, \nu(\gamma_u p_1 \textcolor{blue}{(1 + \frac{1}{ C_t})}, \varepsilon, \delta) \big)$ \Comment{Scaled Gaussian noise}
\State $\bar{\phi} \gets  F\big(\{( \mu_i, t_i )\}_{i=1}^{C_t}\big) + \textcolor{blue}{p_1 \times G_\theta\big(\{( \mu_i, t_i )\}_{i=1}^{C_t}\big)} + \xi$ \Comment{Noisy prediction}
\State $\phi \gets \min(\gamma_u, \max(0, \bar{\phi} ))$ \Comment{Clip output by public knowledge}
\end{algorithmic}
 \label{algo:mlp_lipschitz}
\end{algorithm}

\begin{lemm} \label{lemm:algo1_dp}
    Algorithm~\ref{algo:mlp_lipschitz} satisfies $(\varepsilon,\delta)$-DP
\end{lemm}
\textit{Proof: } Function $G_\theta(\cdot)$ has Lipschitz constant with respect to an individual message $\frac{1}{C_t}$ per Theorem~\ref{thm:lip_mlp}.
When the message values are clipped and the timesteps $t_i$ are fixed by assumption,  $G_\theta(\cdot)$ has sensitivity $\frac{\gamma_u}{C_t}$ w.r.t. each message value $\mu_i$.
Combining this sensitivity with the sensitivity of FN at $p_1\gamma_u$, c.f. Theorem~\ref{thm:sensitivity_14day}, the sensitivity of the composed neural augmentation is $\gamma_u p_1  (1 + \frac{1}{C_t})$.
We add noise according to the Analytical Gaussian Mechanism with this sensitivity, using $\nu(\gamma_u p_1 (1+\frac{1}{C_t}),\varepsilon, \delta)$ to determine the noise variance, ensuring $(\varepsilon,\delta)$-DP.\hfill $\square$

The function $\nu(\Delta,\varepsilon, \delta)$ implements Algorithm 1 in~\citet{balle2018gauss} to compute the minimum noise variance for the Analytical Gaussian Mechanism.

\textbf{Learning under Lipschitz constraints } To learn a Lipschitz-constrained neural network with stochastic gradient descent, we use the Power Iteration method to approximate the spectral norm of linear layers at each step during training~\citep{miyato2018spectral}. The Power Iteration method, however, only approximates the spectral norm. Therefore, after training, we calculate the singular values exactly and project the weights accordingly. In practice, we find that the spectral norms after training are close to 1, and the projection step has negligible influence on the final performance.

\subsection{Experimental details} \label{sec:main_exp_details}
The functions $g^{(1)}_{\theta_1}(\cdot)$ and  $g^{(2)}_{\theta_2}(\cdot)$ are each a multilayer perceptron of $M=8$ layers of width $w=64$. 
 
Data for training with the neural augmentation module is obtained from running the simulator three times for 100 time steps, once each time for the training, validation, and test set. Due to a large imbalance, negative samples are subsampled at random to match the number of positive samples. In the simulation, users get tested according to the estimated \covidscore\ and self isolate upon a positive test. In this way, better predictions potentially lead to a lower overall virus spread. Further details are described in Section~\ref{app:experimental_details}. In the experimental results, we report the mean and 90\% confidence interval of the mean after ten random restarts. The randomness arises from variability in contact patterns in the simulator, stochastic disease dynamics, and the additive noise required for DP.

\begin{figure}
\centering
\begin{minipage}{.48\textwidth}
  \centering
  \includegraphics[width=.99\linewidth]{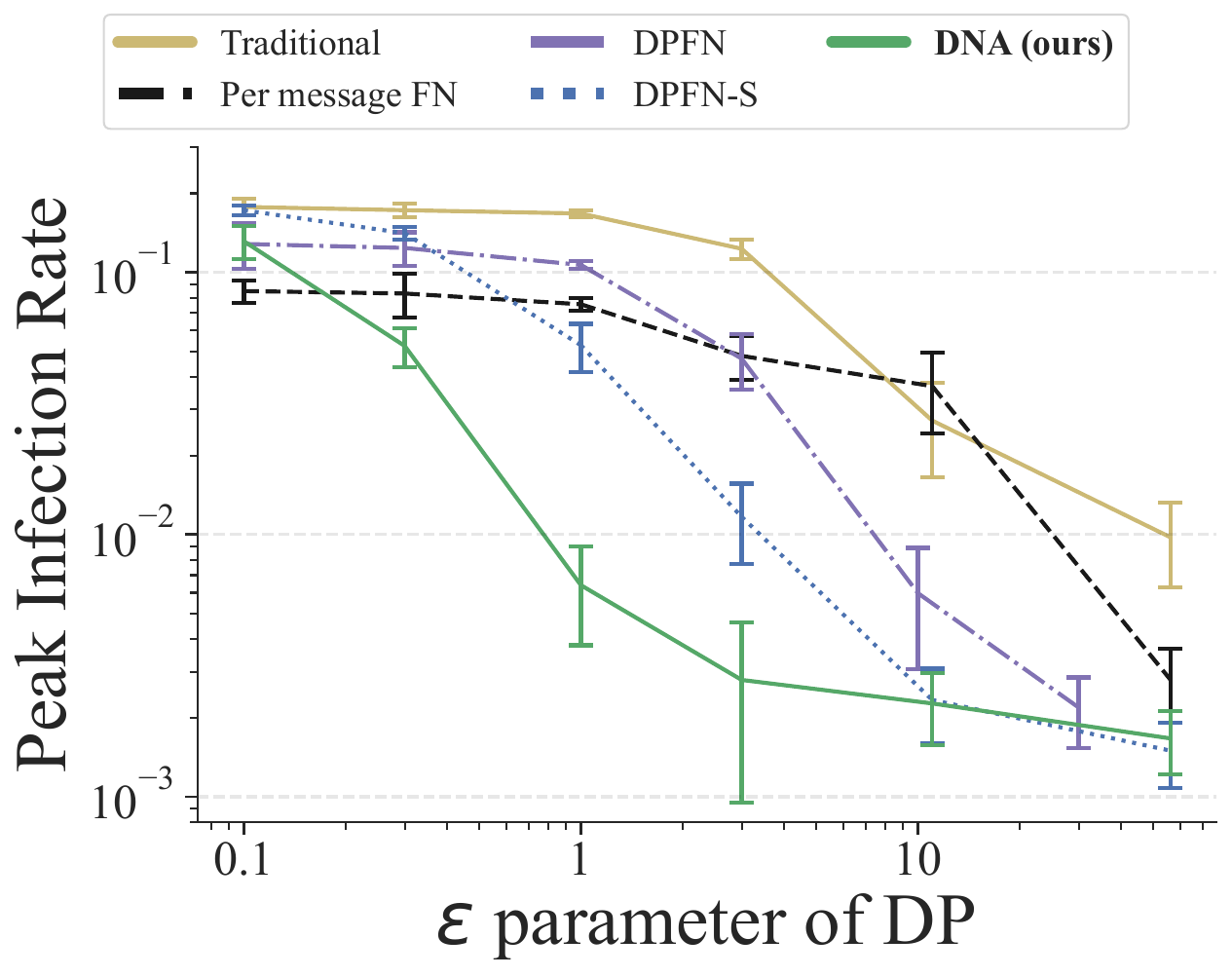}
  \captionof{figure}{The trade-off between peak infection rate (y-axis) and the $\varepsilon$ DP parameter (x-axis). At the crucial setting of $\varepsilon=1$, our method, DNA, achieves a significantly lower peak infection rate.}
  \label{fig:dpgnn_p1sens}
\end{minipage}%
\ \ \ \ \ \ 
\begin{minipage}{.48\textwidth}
  \centering
    \centering
    \setlength{\tabcolsep}{4pt}
\begin{tabular}{c | D{, }{_\pm}{-1} D{, }{_\pm}{-1} }
& \multicolumn{1}{c}{\textbf{DPFN-S} (\permil)}
& \multicolumn{1}{c}{\textbf{DNA} (\permil)}\\[1mm]
\hline   \rule{0pt}{2.4ex}
\textit{Follow protocol \ \ \ \ \ } & & \\[.6mm]
100\%   & 52.7,_{10.9} & 6.4,_{2.6} \\
80\%    & 60.4,_{9.6} & 6.4,_{2.2} \\
50\%    & 100.1,_{4.4} & 27.2,_{8.6} \\[1mm]
      \hline \rule{0pt}{2.4ex}
      \textit{Noisy tests \ \ \ \ \ } & & \\[.6mm]
FPR 1\%, FNR .1\%  & 52.7,_{10.9} & 6.4,_{2.6} \\
FPR 10\%, FNR 1\%  & 81.3,_{2.6} & 19.5,_{2.5} \\
FPR 25\%, FNR 3\% & 130.4,_{1.5} & 81.3,_{1.8} \\[1mm]
\end{tabular}
  \captionof{table}{Two ablation experiments to test the robustness. When up to 50\% of the agents don't follow the protocol, or when the tests become more noisy, the DNA method still achieves lower PIR than the method without neural augmentation. Numbers are in one-per-thousand (\permil).}
  \label{tab:robustness}
\end{minipage}
\end{figure}

\vspace{-2mm}
\section{Experimental results}

We evaluate the four methods on a simulation of 10.000 agents for 100 days and report the peak infection rate (PIR), which is the largest fraction of simultaneously infected individuals. A high number corresponds to a large pandemic, which strains the healthcare system and society with all its consequences. The parameters of the neural augmentation module are learned on a dataset based on the simulator, where we measure the Area Under the Receiver-Operator Curve (AUC) as a performance metric. On the test set, predictions from FN achieve an AUC of 77.0, while the combination with the neural augmentation module achieves 83.1 AUC. This improvement of more than six points in AUC already shows the benefits of neural augmentation on the prediction task.

Figure~\ref{fig:dpgnn_p1sens} shows the results of deploying the methods on the \openabm\ simulator. We compare neural augmentation against traditional contact tracing, which counts only the number of positive contacts (Traditional, ~\citet{baker2021epidemic}), against DPFN with privacy analysis per message (level 1, per-message FN), against privacy analysis per day (level 2, DPFN), and against privacy analysis by sensitivity (level 3, DPFN-S). The results show a trade-off between infection rates and privacy. For $\varepsilon=0.1$, which is considered a very strict DP, all methods have a large PIR. On the other side, for $\varepsilon=30$, which is not considered private, most methods have a low PIR. Expert studies identify $\varepsilon=1$ as a target for DP~\citep{choosing_epsilon,choosing_epsilon_02} and at this $\varepsilon=1.0$-DP, the experimental results show that our DNA method has significantly lower PIR than other methods.\footnote{Note from July 2026: some plotted points are shifted from the nominal x-axis values ($1.0, 3.0, \ldots$) because we originally used the Gaussian variance formula from~\citet{dwork_dp} for $\varepsilon>1$, which is incorrect; the $\varepsilon$ values were recalculated using the Analytical Gaussian Mechanism of~\citet{balle2018gauss}.}

We run two ablation experiments to test the robustness of our method. First, we simulate that agents don't adhere to the protocol of voluntarily isolating after a positive test, e.g. they continue to interact with other agents. Up to 50 \% non-adherence, the DNA method gets significantly lower PIR. Secondly, we run the simulation with higher false positive and false negative rates. The results in Table~\ref{tab:robustness} show that even with tests as noisy as 25\% FPR and 3\% FNR, DNA achieves significantly lower PIR. 

\section{Discussion and conclusion}

We propose a novel algorithm for statistical contact tracing using neural augmentation, which can learn patterns from data that are not captured by a statistical model. The algorithm is decentralized and maintains differential privacy against a recently identified attack scenario on contact tracing.

Further research is needed in two directions. Our algorithm quantifies the DP per message and repeated contacts are treated as individual messages, but a group of attackers could yield more information by repeatedly contacting the same individual. More research is needed on privacy composition in iterative decentralized inference methods. Secondly, in the classification setting, DP is known to exacerbate biases with respect to minority groups~\citep{farrand2020neither}, and although decentralized contact tracing is a different context, this effect should be further investigated.

The early consequences of a pandemic like \covidn\ can be mitigated with contact tracing. We improve the predictions of a differentially private algorithm, and experiments show that our method significantly reduces the peak infection rate, especially at the level of $\varepsilon=1$ per message. This is a crucial step for differentially private and decentralized contact tracing in case a new pandemic arises.

\bibliography{iclr2024_conference}
\bibliographystyle{iclr2024_conference}

\appendix
\newpage 
\section{Appendix}
\section*{Acknowledgements}
This work is financially supported by Qualcomm Technologies Inc., the University of Amsterdam and the allowance Top consortia for Knowledge and Innovation (TKIs) from the Netherlands Ministry of Economic Affairs and Climate Policy. Qualcomm AI research is an initiative of Qualcomm Technologies, Inc. and/or its subsidiaries. Correspondence may go to \href{mailto:r.romijnders@uva.nl}{r.romijnders@uva.nl}.

\subsection{Notation} \label{app:notation}

We repeat the most important notation in this paper.
\begin{itemize}
    \item $\phi_{u,t}$ is the \covidscore\ of user $u$ on day $t$.
    \item $\mu_i$ is a message sent in decentralized inference. This is a function of the \covidscore\ of the sender, $\phi_{c,t}$.
    \item $z_{u,t}$ is the random variable of a user $u$ on day $t$ and takes on the values $\{S, E, I, R\}$ for the Susceptible, Exposed, Infected, Recovered states.
    \item $z_{u,1:T} =  \{z_{u,1}, z_{u,2}, \cdots, z_{u,T} \}$ corresponds to the set of random variables for the time range 1 up to and including $T$.
    \item $C$ indicates the number of contacts; $C_t$ is the number of contacts on day $t$.
    \item $p_0, p_1, g, h \ \in (0, 1)$ are scalar model parameters. Its values are taken from previous studies such as~\citep{crisp,romijnders2023notimetowaste}.
    \item $b(z)$: the functions $b$ generally indicate beliefs~\citep{DBLP:conf/uai/Rosen-ZviJY05}, which could be either $b_u(z_{u,t})$ for a particular user or $b_{N(u)}(z_{N(u)})$ for a set of users.
    \item $N(u,t)$ are the neighbors of user $u$ at day $t$. We write $N(u)$ and $N(t)$ for briefness when $u$ or $t$  is clear from context.
    \item $\varepsilon > 0, \ \delta \in [0,1)$ are the parameters for differential privacy.
    \item $\gamma_u$ is the clipping bound for messages between decentralized agents, e.g. $\mu_i \in [0, \gamma_u]$.
\end{itemize}

\subsection{Background}\label{app:background}

\textbf{Details on the statistical model: } The random variables are written as $z_{u,t}$ for user $u$, at time step $t$. For a particular user, the variables $z_{u,t}, z_{u,t+1}, \cdots$ form a Markov chain. The Markov chain is described by a conditional distribution between one timestep and the next:

\begin{alignat}{2}\label{eqn:seir_dynamics_full}
  P(z_{u,t+1}|z_{u,t}, z_{N(u,t)})  = \begin{cases}
    \psi(u, t, z_{N(u,t)})    &  S \rightarrow S\\
    1-\psi(u, t, z_{N(u,t)})  &  S \rightarrow E\\
    1-g                       &  E \rightarrow E\\
    g                         &  E \rightarrow I\\
    1-h                       &  I \rightarrow I\\
    h                         &  I \rightarrow R\\
    1                         &  R \rightarrow R\\
    0                         & \mbox{otherwise}
    \end{cases} \,
\end{alignat}

The prior for the first timestep for every user is: 
\begin{alignat}{2}\label{eqn:seir_prior}
  P(z_{u,1})  = \begin{cases}
    1-p_0    &  z_{u,1} = S \\
    p_0      & z_{u,1} = E \\ 
    0        & \mbox{otherwise.}
    \end{cases} \,
\end{alignat}

The function $\psi(\cdot)$ is the statistical model for virus transmission over a user contact. This dynamic is described with a noisy-OR model, c.f. Equation~\ref{eqn:crisp_noisy_or} and repeated here:~
\begin{align}\label{eqn:crisp_noisy_or__psi}
   \psi(u, t, z_{N(u,t)}) = (1-p_0) (1-p_1)^{|\{z_c \ \in \ z_{N(u, t)}: \ z_{c}=I \}|} .
\end{align}

Further introduced variables are $g$, $h$, $p_0$, and $p_1$, which are set to value from previous literature~\citep{crisp, romijnders2023notimetowaste}. A user can have contact with multiple users on a particular day. As such, $z_{N(u,t)}$ denotes the set of random variables of all contacts of user $u$ at time step $t$. 

Particular users can also take a test for \covidn. These tests can be assigned randomly, or a potential contact tracing algorithm could predict a \covidscore\ per user and prompt users with a high score to take a test. The test may have a false positive or false negative result with respect to the underlying state. We denote the False Positive rate (FPR) by $\beta$ and denote the False Negative rate (FNR) by $\alpha$. The conditional observation model then follows:

\begin{equation}
    P\left(o_{u,t}|z_{u, t}\right) =  
    \begin{cases}
    \alpha & \textrm{if}\ z_{u,t}=I \wedge o_{u,t}=0 \\
    1-\alpha & \textrm{if}\ z_{u,t}=I \wedge o_{u,t}=1 \\
    1-\beta & \textrm{if}\ z_{u,t}\in\left\{ S,E,R\right\} \wedge o_{u,t}=0 \\
    \beta & \textrm{if}\ z_{u,t}\in\left\{ S,E,R\right\} \wedge o_{u,t}=1  
    \end{cases}. \label{eqn:test_outcome}
\end{equation}

\fbox{\begin{minipage}{35em}
\begin{lemm}\label{lemm:fn_repeat} FN update of noisy-or model \\ 

This lemma repeats Equation 8 in~\citet{romijnders2023notimetowaste}. \\ 

The expectation of the noisy-OR model under FN for user $v$ after $C$ contacts happened at timestep $\tau$. The random variables of these contacts are denoted $z_{c, \tau}$. This has the corresponding FN belief $b_c(z_{c, \tau})$, with message parameter $\mu_{c, \tau}$. When clear from context, we write the message parameter as $\mu_i$ for contact $i=c$. A central assumption in FN is that the belief over the set of neighbor nodes, $B_{N(u)}$, follows a factor distribution $B_{N(u)} = \prod_{c=1}^C b_c(z_{c, \tau})$.

\begin{align} \nonumber
    \mathbb{E}_{B_{N(u)}(z_{N(u)})}&\Bigr[ p(z_{v,\tau+1}=S|z_{v,\tau}=S,z_{N(v, \tau)}=\{z_{c,\tau}\}_{c=1}^{C})\Bigr]\  \\ \nonumber
    &= E_{B_{N(u)}(z_{N(u)})}\left[(1-p_0)\prod_{c=1}^{C}\textcolor{black}{(1-p_1)^{\mathbf{1}[z_{c,\tau}]}}\right] \\ 
    &= (1-p_0)\prod_{c=1}^{C}  E_{b_c(z_{c,\tau})}\left[(1-p_1)^{\mathbf{1}[z_{c,\tau}]}\right] \nonumber \\ 
    &= (1-p_0)\prod_{c=1}^{C}  \left[1-p_1 \mu_{c, \tau} \right]  \label{eqn:crisp_dynamics_under_exp}
\end{align}

Notation $\mathbf{1}[\cdot]$ indicates the Iverson bracket for having state I, which evaluates to 1 when the argument has state I and 0 otherwise.

\end{lemm}
\end{minipage}}

\subsection{Global sensitivity proof} \label{app:global_sens}

This section provides the proof for Theorem~\ref{thm:sensitivity_14day}. We aim to bound the sensitivity of the FN algorithm on two adjacent datasets, as defined in Equation~\ref{eqn:sensitivity_definition}. The main establishment of this proof is to address the sensitivity with respect to the value of one message in the context of arbitrarily many other statistical messages on that day and  other days. 

\begin{figure}[t]
    \centering
    \includegraphics[width=0.7\textwidth]{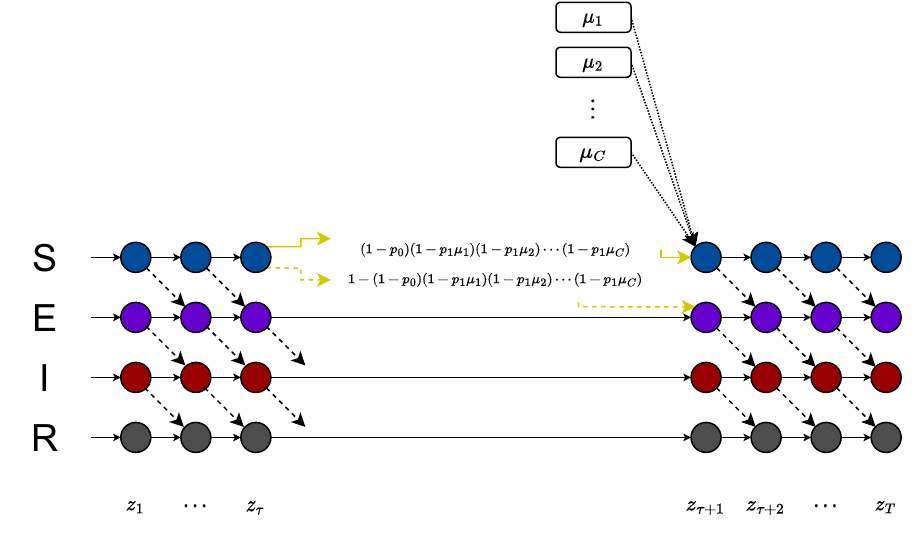}
    \caption{This diagram illustrates the proof setup in Section~\ref{app:global_sens}. The value of $\mu_1$ on day $\tau$ occurs in the conditional probability distribution $p(z_{\tau+1} | z_{\tau} = S, z_{N(u)})$. Filled lines indicate a transition to an equal state, and dashed lines indicate a transition to the next state in the $S \rightarrow E \rightarrow I \rightarrow R$ order.}
    \label{fig:global_sens_proof}
\end{figure}

Each dataset has all contacts' messages in a time window of the past $T$ days. Figure~\ref{fig:global_sens_proof} provides a schematic illustration of the proof. Without loss of generality, we consider two adjacent datasets where the value of a message, $\mu_{c, \tau}$, of contact $c$ on day $\tau$ from dataset $D$ to dataset $D'$ is changed. On that day, assume there are $C$ contacts in total, $c=1, c=2, \cdots, c=C$. Each contact sends a message $(1-p_1 \mu_{1, \tau}), (1-p_1 \mu_{2, \tau}), \cdots, (1-p_1 \mu_{C, \tau})$. Here $p_1 \in (0, 1)$ is a model parameter. The definition of these messages follow from~\citep{romijnders2023notimetowaste}, and are repeated in Lemma~\ref{lemm:fn_repeat}.

We introduce two shorthand notations for a set of random variables. We write the set of random variables in a sequence of days $z_{1:T} = \{z_1, z_2, \cdots, z_T \}$. We also use $z_{N(u, 1:T)}$ to denote the neighboring nodes, contacts, of user $u$ in timesteps 1 up to and including $T$, where a neighbor could be any user $v$ not equal to $u$. We write $Z_{N(u)}$ when the timesteps are evident from the context.

The general update rule for FN is:
\begin{align}
     F(D) &= b(z_T = I)  \nonumber \\ 
    &= \sum_{z_{1:T-1}} \mathbb{E}_{b_{N(u)}}[ p(z_T=I, z_{1:T-1} | z_{N(u, 1:T)}) ] \\ 
    &= \sum_{z_{1:T-1}} \mathbb{E}[p(z_T=I, z_{\tau+1:T-1} | z_{1:\tau}, z_{N(u, \tau+1:T)})] \mathbb{E}[p(z_{1:\tau} |  z_{N(u, 1:\tau)})] \label{eqn:bigpresummation} \\
    &= \sum_{z_{1:T-1}} \mathbb{E}[p(z_T=I, z_{\tau+1:T-1} | z_{\tau}, z_{N(u, \tau+1:T)})] \mathbb{E}[p(z_{1:\tau} |  z_{N(u, 1:\tau)})]  \label{eqn:bigsummation}
\end{align}

The equalities in Equations~\ref{eqn:bigpresummation} and~\ref{eqn:bigsummation} follow from the Markov property of the model, c.f. Equation~\ref{eqn:seir_dynamics_full}. The Markov property implies the conditional independence:~
\begin{align} \label{eqn:cond_indep}
  z_{\tau+1:T} \ \bot \ z_{1:\tau-1} \ &| \ z_\tau  \\
    z_{\tau+1:T} \ \bot \ z_{N(u, 1:\tau)} \ &| \ z_\tau  .
\end{align} 

We split summation in Equation~\ref{eqn:bigsummation} in three parts: timesteps before $\tau$, the transition at timestep $\tau$, and all timesteps after $\tau$.
In the following, all expectations are taken with respect to the factored FN beliefs, so we use shorthand $\mathbb{E}[\cdot]$ to indicate $\mathbb{E}_{b_{N(u)}}[\cdot]$, c.f. Lemma~\ref{lemm:fn_repeat}. 

\begin{align}
 F(D) &=  \sum_{z_{\tau+1:T-1}} \sum_{z_\tau} \mathbb{E}[p(z_T=I, z_{\tau+1:T-1} | z_{\tau}, z_{N(u, \tau+1:T)})] \sum_{z_{1:\tau-1}} \mathbb{E}[p(z_{1:\tau} |  z_{N(u, 1:\tau)})] \\ 
     &= \sum_{z_{\tau+1:T-1}} \sum_{z_\tau} \mathbb{E}[p(z_T=I, z_{\tau+2:T-1} | z_{\tau+1}, z_{N(u, \tau+2:T)})\textcolor{blue}{p(z_{\tau+1} | z_{\tau}, z_{N(u, \tau+1)})}]   \nonumber \\ 
     & \qquad \cdot \sum_{z_{1:\tau-1}} \mathbb{E}[p(z_{1:\tau} |  z_{N(u, 1:\tau)})]. \label{eqn:fn_three_sums_blue}
\end{align}

The conditional distribution for $z_{\tau+1}$ in Equation~\ref{eqn:fn_three_sums_blue} is highlighted in blue for clarity, as we will consider two explicit cases for $z_\tau$ in this conditional distribution for the following analysis. The random variable $z_\tau$ can be either in state $S$ or not in state $S$, which are states $\{ E, I, R \}$. From state $S$ the only non-zero conditional probabilities, under Equation~\ref{eqn:seir_dynamics_full}, are to state $S$ and state $E$ for $z_{\tau+1}$.
For brevity of notation, we will write $z_{N(t_1:t_2)}$ for $z_{N(u, t_1:t_2)}$ when the user is clear from context.

\small
\begin{alignat}{3}
F(D) = & \nonumber \\
      \Big( \sum_{z_{\tau+2:T-1}} \big( &\mathbb{E}[p(z_T=I, z_{\tau+2:T-1} | z_{\tau+1} = S, z_{N( \tau+2:T)})]  \mathbb{E}[p(z_{\tau+1} = S | z_{\tau} =S, z_{N( \tau+1)}) ]   \nonumber \\
       +&\mathbb{E}[p(z_T=I, z_{\tau+2:T-1} | z_{\tau+1} = E, z_{N( \tau+2:T)})]  \mathbb{E}[p(z_{\tau+1} = E | z_{\tau} =S, z_{N( \tau+1)}) ] \big)  \nonumber \\ 
       & \qquad \cdot \sum_{z_{1:\tau-1}}  \mathbb{E}[p(z_\tau = S, z_{1:\tau-1} |  z_{N( 1:\tau)})] \Big) \nonumber \\
       +& \Big(\sum_{z_{\tau+1:T-1}} \sum_{z_\tau \in \{ E, I, R \} } \mathbb{E}[p(z_T=I, z_{\tau+1:T-1} | z_\tau \neq S , z_{N( \tau+2:T)}) ]  \sum_{z_{1:\tau-1}}  \mathbb{E}[p(z_\tau, z_{1:\tau-1} |  z_{N( 1:\tau)})] \Big) \label{eqn:expansion_tau}.
\end{alignat}
\normalsize

The outer sum in Equation~\ref{eqn:expansion_tau} starts at $\tau+2$ because for timestep $\tau+1$ in the first factor, the states $S$ and $E$ are explicitly written. 
 When the contact occurs on day $\tau$ and we name this contact user $v$, then $z_{v,\tau} \in z_{N(u,\tau+1)}$. In Equation~\ref{eqn:expansion_tau}, conditioned on $z_\tau \neq S$, the random variables $z_T=I, z_{\tau+1:T-1}$ are conditionally independent from $z_{N(u, \tau+1)}$. Therefore, we define two factors that are constant w.r.t. the random variables of the contacts on day $\tau$:

\begin{align}
    K_0 &=  \sum_{z_{\tau+1:T-1}} \sum_{z_\tau \in \{ E, I, R \} } \mathbb{E}[p(z_T=I, z_{\tau+1:T-1} | z_\tau  , z_{N(\tau+2:T)}) ]  \sum_{z_{1:\tau-1}}  \mathbb{E}[p(z_\tau, z_{1:\tau-1} |  z_{N(1:\tau)})] \label{eqn:k0_def} \\ 
    K_1 &= \sum_{z_{1:\tau-1}}  \mathbb{E}[p(z_\tau = S, z_{1:\tau-1} |  z_{N(1:\tau)})] \label{eqn:k1_def}.
\end{align}

Rewrite Equation~\ref{eqn:expansion_tau} and fill in the conditional distribution as defined in Equation~\ref{eqn:crisp_dynamics_under_exp}. For ease of notation, all messages on day $\tau$ will be written as $\mu_i$ instead of $\mu_{i,\tau}$.~
\small
\begin{alignat}{3}
F(D) &=  K_0 + K_1 \sum_{z_{\tau+2:T-1}} \Big( \mathbb{E}[p(z_T=I, z_{\tau+2:T-1} | z_{\tau+1} = S, z_{N(\tau+2:T)})]  \mathbb{E}[p(z_{\tau+1} = S | z_{\tau} =S, z_{N(\tau+1)}) ]   \nonumber \\
       & \qquad +\mathbb{E}[p(z_T=I, z_{\tau+2:T-1} | z_{\tau+1} = E, z_{N(\tau+2:T)})]  \mathbb{E}[p(z_{\tau+1} = E | z_{\tau} =S, z_{N(\tau+1)}) ] \Big)  \nonumber \\ 
        &=  K_0 + K_1 \Big(  \mathbb{E}[p(z_{\tau+1} = S | z_{\tau} =S, z_{N(\tau+1)}) ]\sum_{z_{\tau+2:T-1}}  \mathbb{E}[p(z_T=I, z_{\tau+2:T-1} | z_{\tau+1} = S, z_{N(\tau+2:T)})]     \nonumber \\
       & \qquad + \mathbb{E}[p(z_{\tau+1} = E | z_{\tau} =S, z_{N(\tau+1)}) ] \sum_{z_{\tau+2:T-1}} \mathbb{E}[p(z_T=I, z_{\tau+2:T-1} | z_{\tau+1} = E, z_{N(\tau+2:T)})]   \Big) \nonumber \\ 
       &= K_0 + K_1 \Big( (1-p_0)(1-p_1 \mu_{1})(1-p_1 \mu_{2})\cdots (1-p_1 \mu_{C})     \nonumber \\
       & \qquad \qquad \qquad  \cdot \sum_{z_{\tau+2:T-1}}  \mathbb{E}[p(z_T=I, z_{\tau+2:T-1} | z_{\tau+1} = S, z_{N(\tau+2:T)})] \nonumber \\ 
       & \qquad \qquad +(1-(1-p_0)(1-p_1 \mu_{1})(1-p_1 \mu_{2})\cdots (1-p_1 \mu_{C}))   \nonumber \\ 
       & \qquad \qquad \qquad \cdot \sum_{z_{\tau+2:T-1}}  \mathbb{E}[p(z_T=I, z_{\tau+2:T-1} | z_{\tau+1} = E, z_{N(\tau+2:T)})] \Big) \label{eqn:fn_before_keks}
\end{alignat}
\normalsize

We further introduce two constants w.r.t. the message value $\mu_i$:

\begin{align}
    K_S &= \sum_{z_{\tau+2:T-1}} \mathbb{E}[p(z_T=I, z_{\tau+2:T-1} | z_{\tau+1} = S, z_{N(\tau+2:T)})] \label{eqn:ks_def}   \\ 
    K_E &= \sum_{z_{\tau+2:T-1}} \mathbb{E}[p(z_T=I, z_{\tau+2:T-1} | z_{\tau+1} = E, z_{N(\tau+2:T)})]     \label{eqn:ke_def}  
\end{align}

\fbox{\begin{minipage}{35em}
\begin{lemm}\label{lemm:constants} Constants $K_0$, $K_1$, $K_S$, and $K_E$ are in $[0, 1)$. \\ 

The conditional probability distributions in Equations~\ref{eqn:seir_dynamics_full} and ~\ref{eqn:crisp_noisy_or} take value in $[0, 1)$ for any variable realization because the model parameters are chosen such that $ 0 < p_0, p_1, g, h < 1$ and the joint probability distributions are on a discrete domain. This holds for all conditional distributions in the definitions of $K_0$, $K_1$, $K_S$, and $K_E$. \\

The expectation operator, $\mathbb{E}[\cdot]$, is a convex combination, and a convex combination of numbers in $[0, 1)$, is itself in $[0, 1)$. The constants $K_0$, $K_1$, $K_S$, and $K_E$ in Equations~\ref{eqn:k0_def}, ~\ref{eqn:k1_def}, ~\ref{eqn:ks_def}, ~\ref{eqn:ke_def} can all be rewritten to an expectation of a joint probability distribution and are, therefore, in $[0, 1)$.

\end{lemm}
\end{minipage}}

Now, we can rewrite Equation~\ref{eqn:fn_before_keks}, replacing all factors that are constant with respect to $\mu_1$.

\begin{align}
    F(D) =  \big(&K_S (1-p_0)(1-p_1 \mu_{1})(1-p_1 \mu_{2})\cdots (1-p_1 \mu_{C})   \nonumber \\
       &+ K_E (1-(1-p_0)(1-p_1 \mu_{1})(1-p_1 \mu_{2})\cdots (1-p_1 \mu_{C})) \big) K_1 + K_0. \label{eqn:k_patterns}
\end{align}

Finally, we obtain the sensitivity from taking the difference between the two adjacent datasets $D,D'$. The FN inference update equation is invariant to a permutation of the messages in the same day. Therefore, as mentioned, without loss of generality, we assume that the change from $D$ to $D'$ happens in contact $1$, which changes its message value from $\mu_1$ to $\mu'_1$.  

\begin{align}
  \Vert F(D) - F(D') \Vert  &=  \Vert   \Big( \big(K_S (1-p_0)(1-p_1 \mu_{1})(1-p_1 \mu_{2})\cdots (1-p_1 \mu_{C})  \nonumber \\
   & \qquad \  + K_E (1-(1-p_0)(1-p_1 \mu_{1})(1-p_1 \mu_{2})\cdots (1-p_1 \mu_{C})) \big) K_1 + K_0 \Big)- \nonumber \\
   & \qquad  \Big( \big(K_S (1-p_0)(1-p_1 \mu'_{1})(1-p_1 \mu_{2})\cdots (1-p_1 \mu_{C}) \nonumber \\
   & \qquad \  + K_E (1-(1-p_0)(1-p_1 \mu'_{1})(1-p_1 \mu_{2})\cdots (1-p_1 \mu_{C})) \big) K_1 + K_0 \Big) \Vert  \\ 
   & \leq \Vert  \Big( K_S (1-p_0)(1-p_1 \mu_{1})(1-p_1 \mu_{2})\cdots (1-p_1 \mu_{C})  \nonumber \\
   & \qquad \ + K_E (1-(1-p_0)(1-p_1 \mu_{1})(1-p_1 \mu_{2})\cdots (1-p_1 \mu_{C}))  \Big)- \nonumber \\
   & \qquad  \Big( K_S (1-p_0)(1-p_1 \mu'_{1})(1-p_1 \mu_{2})\cdots (1-p_1 \mu_{C}) \nonumber \\
   & \qquad  \ + K_E (1-(1-p_0)(1-p_1 \mu'_{1})(1-p_1 \mu_{2})\cdots (1-p_1 \mu_{C})) \Big) \Vert  \label{eqn:dropk0k1} \\ 
   & \leq \Vert  (1-p_1 \mu_{1}) - (1-p_1 \mu'_{1}) \Vert  \label{eqn:dropkeks} \\ 
   & = p_1 \Vert \mu_{1} - \mu'_{1}\Vert  
\end{align}

Arriving at Equation~\ref{eqn:dropk0k1}, we use Lemma~\ref{lemm:constants} that $0 \leq K_0 < 1$ and  $0 \leq K_1 < 1$.
Likewise, arriving at Equation~\ref{eqn:dropkeks}, we use that $0 \leq K_S < 1$ and  $0 \leq K_E < 1$. 

If we assume that every message $\mu_i$ is clipped to the range $[0, \gamma_u]$, then we prove the sensitivity. 

\begin{equation}
 \Delta =\max_{\mu_1, \mu'_1  \in [0, \gamma_u]} \norm{F\big( \ \{(\mu_1, t_1)\} \cup D \ \big) - F\big( \ \{(\mu'_1, t_1)\} \cup D \ \big)} \leq p_1 \gamma_u \qquad \forall \ D
\end{equation}

\newpage
\subsection{Experimental details} \label{app:experimental_details}

The different methods are compared on a simulator for \covidn. We need a simulator to evaluate the interaction between predictions from each method, how that affects which individuals get tested, and the resulting change in infection rates if infectious individuals go in isolation. All methods in this paper are decentralized to ensure the locality of data. Each day, the method assigns a \covidscore\ for the local user. The top 8\% of agents with the highest score receive a signal to get tested. An agent that tests positively isolates themselves for ten days, and no contacts occur during this period. We assume that a private algorithm exists to determine the agents with the highest score~\citep{encrypted_top_01,encrypted_top_02}. The tests have a False Positive Rate of 1\% and False Negative Rate of 0.1\%. In Table~\ref{tab:robustness}, we report on two ablation experiments where users do not adhere to the isolation protocol and when the false positive and false negative rates increase.

The settings for the \openabm\ simulator~\citep{openabm} follow the parameter settings in previous work~\citep{baker2021epidemic,romijnders2024protectyourscore}. The simulator has over 150 modifiable parameters and uses different network properties to model agents. The parameters are calibrated against population data from the UK, in terms of age distributions and household and occupation patterns. Due to computational limits, the simulator's experimental results are each with 10.000 agents. The noise rates in Table~\ref{tab:robustness} also follow previous work~\citep{romijnders2024protectyourscore}. The largest values of 25\% False Positive Rate and 3\% False Negative Rate correspond to the maximum allowance for an approved \covidn\ test by the European Centre for Disease Control ~\citep{ecdc}.

For training the neural augmentation module, we use a dataset that is extracted from the \openabm\ simulator --  on different random seeds than later testing. We run the simulator on 100.000 agents for 100 time steps. As the model will be used in a decentralized setting, we store, per agent per timestep, the local messages, observations, and the underlying simulator state. The loss function for training is a Mean Squared Error, and the target label is one if the agent has the infected state, I, and zero if the agent has any other state, e.g., S, E, R. We measure the Area Under Receiver-Operator Curve (AUC) as a performance metric to make model decisions. The test set is obtained with the same protocol but a different random seed. On this test set, FN by itself achieves an AUC of 77.0, while the neural augmentation module achieves 83.1. This improvement of almost six points in AUC already shows the benefits of neural augmentation on the static dataset. Note that without neural augmentation, we found that the best Lipschitz-constrained network achieves only 80.3 AUC. 

The parameter vector $\theta = [\theta_1; \theta_2  ]$ is learned with stochastic gradient descent,  ADAM~\citep{adam_kingma}. The model trains for 40 epochs, with weight decay $10^{-9}$ and learning rate 0.002, which decays to 0.0002 by a cosine learning rate decay schedule~\citep{cosine_lr}. The functions $g^{(1)}_{\theta_1}(\cdot)$ and  $g^{(2)}_{\theta_2}(\cdot)$ are each a multilayer perceptron of $M=8$ layers of width $w=64$, as discussed in Section~\ref{sec:main_exp_details}. The activation function is the Rectified Linear Unit~\citep{relu}. During training, we find that using two or more power iteration steps per gradient descent step results in spectral norms close to 1. After training, we calculate the spectral norms exactly and project the weights accordingly.

All inference algorithms in this study use a finite time window of $T=14$ days. This means that after 14 days, a message from a contact is deleted and has no influence on future predictions anymore. The function for FN, $F(\cdot)$, is also stateless. So, any user can withdraw their data at any time, and it will not be used the next day or thereafter. 

Experiments in this paper use $(\varepsilon,\delta)$ differential privacy. We follow the convention and set $\delta$ smaller than one divided by the dataset size~\citep{blanco2022critical, choosing_epsilon, van2023considerations}. The simulators have an average of fifteen contacts daily, and the time window is fourteen days. Therefore, we set $\delta = \frac{1}{1000}$, which is well below the recommended $\frac{1}{14 \times 15}$.

\end{document}